\pdfoutput=1

\documentclass[11pt]{article}

\usepackage{acl}

\usepackage{times}
\usepackage{latexsym}
\usepackage{arydshln}

\usepackage[T1]{fontenc}

\usepackage[utf8]{inputenc}

\newcommand{\mlft}{\textsc{Ml-ft}} 
\newcommand{\langpair}{\textsc{Lang-Pair}}
\newcommand{\langagnostic}{\textsc{Lang-Agnostic}}
\newcommand{\langfamily}{\textsc{Lang-family}}

\newcommand{\comet}{\textsc{Comet}}

\newcommand{\bleu}{\textsc{Bleu}}

\usepackage{microtype}
\usepackage[T1]{fontenc}
\usepackage{listings}
\usepackage{stfloats}
\usepackage{csquotes}
\usepackage{array}
\usepackage{booktabs}

\usepackage[super]{nth}
\usepackage[inline,shortlabels]{enumitem}
\usepackage{multirow}
\usepackage{makecell}
\usepackage{siunitx}
\usepackage{multicol}
\usepackage{graphicx}
\usepackage{subfig}

\usepackage{graphicx}
\usepackage{svg}
\usepackage{float}
\usepackage{ amssymb }

\usepackage{amsmath,amsfonts,amssymb}
\graphicspath{ {images/} }

\usepackage{tablefootnote}

\title{Language-Family Adapters for Low-Resource \\ Multilingual Neural Machine Translation}

\author{ 
	Alexandra Chronopoulou$^{\bigtriangleup}$ \quad 
    Dario Stojanovski$^{\dag\bigtriangledown}$ \quad
	Alexander Fraser$^{\bigtriangleup}$ \quad \\ 
	$^{\bigtriangleup}$Center for Information and Language Processing, LMU Munich, Germany \\
    $^{\bigtriangleup}$Munich Center for Machine Learning, Germany \\ 
 $^{\bigtriangledown}$Microsoft, Belgrade, Serbia \\
	{\tt achron@cis.lmu.de} 
     \\{\tt dstojanovski@microsoft.com, \tt fraser@cis.lmu.de} 
	}

\begin{document}
\maketitle
\begin{abstract}

Large multilingual models trained with self-supervision  
achieve state-of-the-art results in a wide range of natural language processing tasks.
Self-supervised pretrained models are often fine-tuned on parallel data from one or multiple language pairs for machine translation. Multilingual fine-tuning improves performance on low-resource languages but requires modifying the entire model and can be prohibitively expensive. Training a new adapter on each language pair or training a single adapter on all language pairs without updating the pretrained model has been proposed as a parameter-efficient alternative. 
However, the former does not permit any sharing between languages, while the latter shares parameters for all languages and is susceptible to negative interference. 
In this paper, we propose training  \textit{language-family adapters} on top of mBART-50 to facilitate cross-lingual transfer. Our approach outperforms related baselines, yielding higher translation scores on average when translating from English to 17 different low-resource languages.
We also show that language-family adapters provide an effective method to translate to languages unseen during pretraining.

\end{abstract}

\section{Introduction}
\renewcommand*{\thefootnote}{\fnsymbol{footnote}}
\footnotetext{$\dag$\scalebox{0.91}
{Work done prior to joining Microsoft}}
Recent work in multilingual
natural language processing (NLP) has created models that reach competitive performance, while incorporating 
many languages into a single architecture \cite{devlin-etal-2019-bert, conneau-etal-2020-unsupervised}.
Because of its ability to share cross-lingual representations,
which 
largely benefits lower-resource languages, multilingual neural machine translation (NMT) is an attractive research field \cite{firat-etal-2016-multi, zoph-etal-2016-transfer, johnson-etal-2017-googles,ha-toward-multilingual,zhang-etal-2020-improving,m2m100}. Multilingual models are also appealing because they are more efficient in terms of the number of model parameters, enabling simple deployment \cite{arivazhagan2019massively,aharoni-etal-2019-massively}.
Massively multilingual pretrained models can be used for multilingual NMT, if they are fine-tuned in a \textit{many-to-one} 
 (to map any of the source languages into a  target language, which is usually English) or \textit{one-to-many} (to translate a single source language into multiple target languages) fashion \cite{aharoni-etal-2019-massively,Tang2020MultilingualTW}. Training a \textit{many-to-many} (multiple source to multiple target languages) NMT model \cite{m2m100} has also been proposed.

Multilingual pretrained models generally permit improving translation on low-resource language pairs. Specializing the model to a specific language pair further boosts performance, but is computationally expensive. For example, mBART-50 \cite{Tang2020MultilingualTW}, a model pretrained on monolingual data of 50 languages using denoising auto-encoding with the BART objective \cite{lewis-etal-2020-bart} still has to be fully fine-tuned for NMT. 

To avoid fine-tuning large models, previous work has focused on efficiently building multilingual NMT models. Adapters \cite{rebuffi,houlsby}, which are lightweight feedforward layers added in each Transformer \cite{NIPS2017_3f5ee243} layer, have been proposed as a parameter-efficient fine-tuning method. In machine translation, training a different adapter on each language pair on top of a frozen pretrained multilingual NMT model, has shown to improve results for high-resource languages \cite{bapna-firat-2019-simple}. Low-resource languages do not benefit from this approach though, as adapters are trained with limited data. In a similar vein, \citet{cooper-stickland-etal-2021-recipes} fine-tune a pretrained model for multilingual NMT using a single set of adapters, trained on all languages. Their approach manages to narrow the gap but still does not perform on par with multilingual fine-tuning. 

Many-to-one and one-to-many NMT force languages into a joint space (in the encoder or decoder side) and neglect diversity. One-to-many NMT faces the difficulty of learning a conditional language model and decoding into multiple languages \cite{arivazhagan2019massively, Tang2020MultilingualTW}. To better model target languages, recent approaches propose exploiting both the unique and the shared features \cite{wang-etal-2018-three}, reorganizing parameter-sharing \cite{sachan-neubig-2018-parameter}, decoupling multilingual word encodings \citep{wang2018multilingual}, training NMT models from scratch after creating groups of languages \citep{tan-etal-2019-multilingual}, or inserting language-specific layers \citep{m2m100}. 

In this work, we propose using \textit{language-family} adapters that enable efficient low-resource multilingual NMT. We train  adapters for NMT on top of mBART-50 \citep{Tang2020MultilingualTW}.
 The adapters are trained using bi-text from each language family, while the pretrained model is not updated. Groups of languages are formed based on linguistic knowledge bases. Our approach improves positive cross-lingual transfer, compared to \textit{language-pair adapters} \citep{bapna-firat-2019-simple}, which do not leverage cross-lingual information between languages, and \textit{language-agnostic adapters} \citep{cooper-stickland-etal-2021-recipes}, which are trained on all languages and can suffer from negative interference \citep{wang-etal-2020-negative}. Our approach not only yields better translation scores in the majority of languages examined, but also requires less than 20\% of trainable parameters compared to language-pair adapters, i.e., the most competitive baseline.
 
Our main contributions are:
\begin{enumerate}
    \item A novel, effective approach for low-resource multilingual translation which trains adapters on top of mBART-50 for each language family. In the English-to-many setting which we examine, language-family adapters achieve a $+1$ $\bleu$ improvement over language-pair adapters and $+2.7$ $\bleu$ improvement over language-agnostic adapters on 16 low-resource language pairs from OPUS-100.  
    \item  We propose inserting \textit{embedding-layer adapters} into
the Transformer to encode lexical information and conduct an ablation study to assess their utility.
\item We contrast grouping languages based on linguistic knowledge to grouping them based on the representations of a multilingual pretrained language model (PLM) with a Gaussian Mixture Model (GMM). 
\item  We analyze the effect of our approach when evaluating on languages that are new to mBART-50.
    
\end{enumerate}

\section{Background}

\noindent \textbf{Massively Multilingual Models.} Multilingual masked language models have pushed the 
state-of-the-art on cross-lingual language understanding by training a single model for many languages \citep{devlin-etal-2019-bert,xlm, conneau-etal-2020-unsupervised}. Encoder-decoder Transformer \citep{NIPS2017_3f5ee243} models that are pretrained using monolingual corpora from multiple languages, such as mBART \citep{liu}, outperform strong baselines in medium- and low-resource NMT. 
 mBART-50 \citep{Tang2020MultilingualTW} is an extension of mBART, pretrained in $50$ languages and multilingually fine-tuned for NMT. 
However, while multilingual NMT models are known to outperform strong baselines and  simplify model deployment, they are susceptible to negative interference/transfer \citep{mccann, arivazhagan2019massively, Wang_2019_CVPR, conneau-etal-2020-unsupervised} and catastrophic forgetting \citep{Goodfellow14anempirical} when the parameters are shared across a large number of languages. 
Negative transfer affects the translation quality of high-resource \citep{conneau-etal-2020-unsupervised}, but also low-resource languages \citep{wang-etal-2020-negative}. As a remedy, providing extra capacity to a multilingual model using language-specific modules has been proposed \citep{sachan-neubig-2018-parameter,wang2018multilingual,m2m100,pfeiffer-etal-2022-lifting}. We take a step forward in this direction and train \textit{language-family adapters} on top of a pretrained model. Our approach introduces modular components which leverage the similarities of languages and can better decode into multiple directions, improving results compared to baselines.

\noindent \textbf{Adapters for NMT.} \citet{swietojanski} and \citet{vilar-2018-learning} initially suggested learning additional weights that rescale the hidden units for domain adaptation. Adapter layers \cite{rebuffi,houlsby} are small modules that are typically added to a pretrained Transformer and are fine-tuned on a downstream task, while the pretrained model is frozen.  \citet{bapna-firat-2019-simple} add \textit{language-pair} adapters to a pretrained multilingual NMT model (one set for \textit{each} language pair), to recover performance for high-resource language pairs. \citet{cooper-stickland-etal-2021-recipes} start from an unsupervised pretrained model and train \textit{language-agnostic} adapters (one set for \textit{all} language pairs) for multilingual NMT. \citet{philip-etal-2020-monolingual} train \textit{monolingual} adapters for zero-shot translation, while \citet{ustun2021multilingual} propose \textit{denoising adapters}, i.e., adapters trained using monolingual data, for unsupervised multilingual NMT. \citet{baziotis-hyperadapters} inject language-specific parameters in MNMT using adapters, by generating them from a hyper-network, while \citet{lai-etal-2022-4} adapt a model for both a new domain and a new language pair at the same time by combining domain and language representations using meta-learning with adapters.

We identify some challenges in previous works \cite{bapna-firat-2019-simple,cooper-stickland-etal-2021-recipes}.
Scaling language-agnostic adapters to a large number of languages is problematic, as when they are updated with data from multiple languages, negative transfer occurs. In contrast, language-pair adapters do not face this problem, but at the same time do not allow any sharing between languages, therefore provide poor translation to low-resource language pairs. 
Language-family adapters arguably strike a balance, providing a trade-off between the two approaches, 
and our experiments show that they lead to higher translation quality.

\noindent \textbf{Language Families.} Extensive work on cross-lingual transfer has demonstrated that jointly training a model using similar languages can improve low-resource results in several NLP tasks, such as part-of-speech or morphological tagging \cite{tackstrom-etal-2013-token,straka-etal-2019-udpipe}, entity linking \cite{tsai-roth-2016-cross, rijhwani2018zeroshot}, and machine translation \cite{zoph-etal-2016-transfer, johnson-etal-2017-googles, neubig-hu-2018-rapid, oncevay-etal-2020-bridging}. 
Linguistic knowledge bases \cite{littell-etal-2017-uriel, wals} study language variation and can provide insights to phenomena such as negative interference. 
Languages can be organized together using linguistic information, forming language families.
\citet{tan-etal-2019-multilingual} and \citet{kong-etal-2021-multilingual} leverage  families for multilingual NMT, 
the former by training language-family NMT models from scratch, the latter by training a separate shallow decoder for each family. Instead, our approach keeps a pretrained model frozen and only trains language-family adapters, which is parameter-efficient. Compared to fine-tuning the entire model  ($\mlft$), our approach requires less than $12.5$\% of the trainable parameters, as is shown in Table \ref{table:parameters}.

\section{Language-Family Adapters for Low-Resource NMT}
 
Fine-tuning a pretrained model for multilingual NMT provides a competitive performance, yet is computationally expensive, as all layers of the model need to be updated. 
A parameter-efficient alternative 
is to fine-tune
a pretrained multilingual model for NMT with data from all languages of interest using adapters while keeping the pretrained model unchanged. However, as multiple language representations are encoded in the same parameters, capacity issues arise. Languages are also grouped together, even though they might be different in terms of geographic location, script, syntax, typology, etc. As a result, linguistic diversity is not modeled adequately and translation quality degrades. 

\begin{figure}
	\centering
	\includegraphics[width=1.\columnwidth, page=1]{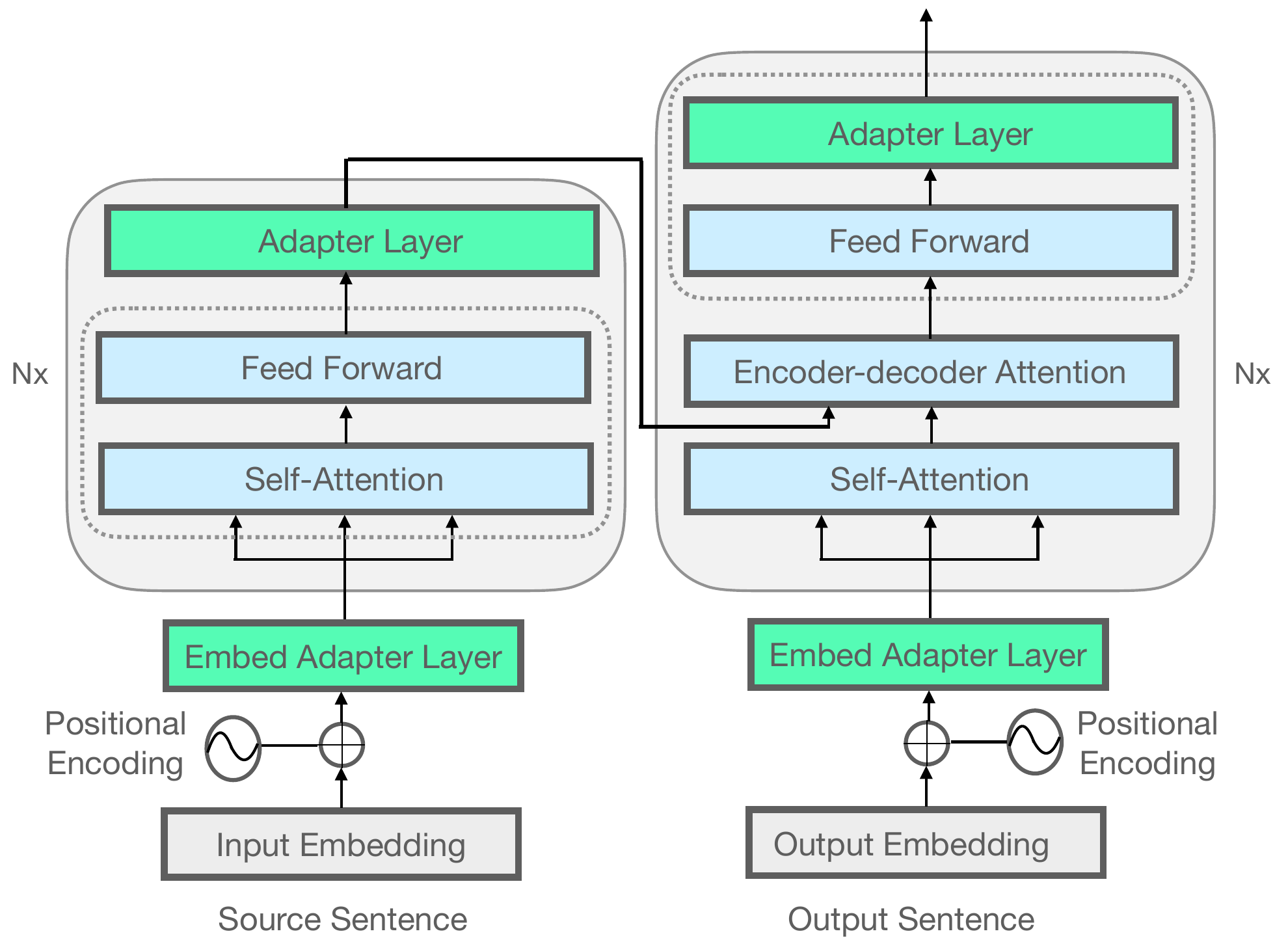}
	\caption{Proposed adapter architecture inside a Transformer model. Adapter layers, shown in green, are trained for NMT. Figure best viewed in color.}
	\label{fig:modelarchitecture}
\end{figure}

We address the limitations of previous methods by proposing language-family adapters for low-resource multilingual NMT. An illustration of our approach is depicted in Figure \ref{fig:modelarchitecture}. We exploit linguistic knowledge to selectively share parameters between related languages and avoid negative interference. 
Our approach is to train adapters using language pairs of a linguistic family on top of a pretrained model, which is not updated.

\subsection{Adapter Architecture}
Adapters are usually added to each Transformer layer. An adapter uses as input the output of the previous layer. Formally:
Let $z_i$ be the output of the $i$-th layer, of dimension $h$. We apply a layer-normalization \cite{ba2016layer}, followed by a down-projection $D$ $\in R^{h\times d}$, a ReLU activation and an up-projection  $U$ $\in R^{d\times h}$, where $d$ is the bottleneck dimension and the only tunable hyperparameter. The up-projection is combined with a residual connection \cite{residual} with $z_i$ according to the following equation: $Adapter_i(z_i) = U \: \text{ReLU}(D \: \text{LN}(z_i)) + z_i$.
This follows \citet{bapna-firat-2019-simple}. Adapters are randomly initialized. 

\subsection{Embedding-layer Adapter}

 Because we keep the token embeddings of mBART-50 frozen, adding flexibility to the model to encode lexical information of the languages of interest is crucial, especially for unseen languages (not part of its pretraining corpus).
Lexical cross-lingual information could be encoded by learning new embeddings for the unseen languages  \cite{artetxe-etal-2020-cross} but this would be computationally expensive. We instead add an adapter after the \textit{embedding} layer, in both the encoder and the decoder, which receives as input the lexical representation of each sequence and aims to capture token-level cross-lingual transformations. 

Our approach draws inspiration from \citet{pfeiffer-etal-2020-mad} and simplifies the invertible adapters structure. We use the large vocabulary of mBART-50 to extend the model to unseen languages.
We note that adding scripts that do not
exist in the vocabulary of mBART-50 is not possible with our approach. We point out that \citet{chronopoulou-etal-2020-reusing, pfeiffer-etal-2021-unks,vernikos-popescu-belis-2021-subword-mapping} have proposed approaches to permit fine-tuning to unseen languages/scripts when using PLMs and we leave further exploration to future work. 

\subsection{Model Architecture}
\label{ssec:architecture}

To train a model for multilingual NMT, we leverage mBART-50, a sequence-to-sequence generative model pretrained on monolingual data from 50 languages using a denoising auto-encoding objective. The model has essentially been trained by trying to predict the original text \textit{X}, given $g(X)$, where $g$ is a noising function that corrupts text. 

We want to fine-tune this model on a variety of language pairs, by leveraging similarities between languages. Our model aims to provide a parameter-efficient alternative to traditional fine-tuning of the entire pretrained model. We note that the pretrained mBART-50 model cannot be used as is for MT, as it has never been trained on the task. 

To this end, we insert adapters after each \textit{feed-forward} layer both in the encoder and in the decoder and we also add embedding-layer adapters. We freeze the pretrained encoder-decoder Transformer and fine-tune \textit{only} the adapters on NMT. We leverage the knowledge of the pretrained model, but encode additional cross-lingual information on each language family using adapters. 
We fine-tune a new set of adapters multilingually on each \textit{language family} and evaluate the performance on and low-resource language pairs.

\section{Experimental Setup}

\noindent \textbf{Data}. We initially fine-tune the model on TED talks \cite{qi-etal-2018-pre}, using data from 17 languages paired to English. We then scale to a larger parallel dataset, using OPUS-100 \cite{zhang-etal-2020-improving} for the same languages paired to English (with the only exception being English-Filipino, which does not appear in OPUS-100). 
For the TED experiments, we choose $17$ languages, $9$ of which 
were
present during pretraining, while $8$ are new to mBART-50. For OPUS-100, we use the same $16$ languages (without Filipino), $9$ of which 
were
present during pretraining and $7$ are new. In both sets of experiments, the languages belong to $3$ language families, namely Balto-Slavic, Austronesian and Indo-Iranian. Balto-Slavic and Indo-Iranian are actually distinct branches of the same language family (Indo-European). The parallel data details are reported in Table \ref{table:data1}. 

\begin{table}[t]
\centering
\small
\resizebox{\columnwidth}{!}{
\begin{tabular}{lrrr}
\toprule

\textbf{Language (code) }& \textbf{Family}   &  \multicolumn{2}{c}{\textbf{Train Set}} \\  
 & & \textbf{TED} & \textbf{OPUS-100} \\ \midrule
$\star$Bulgarian (bg) & BS & 174k & 1M \\
\hspace{1.6mm}Persian (fa) & I & 151k & 1M  \\
$\star$Serbian (sr) & BS  & 137k & 1M  \\
\hspace{1.6mm}Croatian (hr) & BS  & 122k & 1M \\
\hspace{1.6mm}Ukrainian (uk) & BS & 108k & 1M \\
\hspace{1.6mm}Indonesian (id) & A  & 87k & 1M    \\
$\star$Slovak (sk) & BS  & 61k & 1M  \\
\hspace{1.6mm}Macedonian (mk) & BS & 25k & 1M  \\
\hspace{1.6mm}Slovenian (sl) & BS  & 20k & 1M    \\
\hspace{1.6mm}Hindi (hi) & I   & 19k & 534k  \\
\hspace{1.6mm}Marathi (mr) & I & 10k & 27k  \\
$\star$Kurdish (ku) & I & 10k & 45k   \\
$\star$Bosnian (bs) & BS  & 6k & 1M  \\
$\star$Malay (ms) & A & 5k & 1M   \\
\hspace{1.6mm}Bengali (bn) & I &  5k & 1M   \\
$\star$Belarusian (be) & BS & 5k & 67k   \\
$\star$Filipino (fil)  & A & 3k & -    \\

\bottomrule

\end{tabular}}
\caption{Languages used in the experiments. $\star$ indicates languages that are \textit{unseen} from mBART-50, i.e., they do not belong to the pretraining corpus. \textit{BS} stands for Balto-Slavic, \textit{I} for Indo-Iranian, \textit{A} for Austronesian.}

\label{table:data1}
\end{table}
\noindent \textbf{Baselines}. 
We compare the proposed language-family adapters with \textbf{1)}  \textit{language-agnostic} ($\langagnostic$) and \textbf{2)} \textit{language-pair adapters} ($\langpair$). While the adapters are trained using parallel data, mBART-50 (pretrained on monolingual data) is not updated.
 Moreover, we compare our approach to multilingual fine-tuning ($\mlft$), although it requires fine-tuning the entire model and is thus not directly comparable to the parameter-efficient approaches we study. We show this result in the Appendix.

The first baseline, $\langagnostic$ adapters, fine-tunes a set of adapters using data from all languages (similar to \citealp{cooper-stickland-etal-2021-recipes}). The second baseline, $\langpair$ adapters, follows \citet{bapna-firat-2019-simple}: a new set of adapters is trained for each language pair, so no parameters are shared between different language pairs.

\noindent\textbf{Training details}. We start from the mBART-50  checkpoint.\footnote{\url{https://dl.fbaipublicfiles.com/fairseq/models/mbart50/mbart50.pretrained.tar.gz}} We extend its embedding layer with randomly initialized vectors to account for the new languages. We reuse the $250$k sentencepiece \cite{kudo-richardson-2018-sentencepiece} model of mBART-50. We use the fairseq \cite{ott-etal-2019-fairseq} library for all experiments. We select the final models using validation perplexity. If the model is trained on multiple languages (using mixed mini-batches), we use the overall perplexity. We use beam search with size $5$ for decoding and evaluate $\bleu$
scores using Sacre$\bleu$\footnote{Signature ``BLEU+c.mixed+\#.1+s.exp+tok.13a+v.1.5.1''} for OPUS-100 and  Sacre$\bleu$ without tokenization for TED \cite{post-2018-call}.
We also compute $\comet$ \cite{rei-etal-2020-comet} scores using the  \textit{wmt-large-da-estimator-1719} pretrained model. Results are reported in the Appendix.

To train the models, we freeze mBART-50. We fine-tune the $\langfamily$, $\langagnostic$ adapters in a multilingual, one-to-many setup, using English as the source language.
$\langpair$ adapters are fine-tuned for each language pair. All models have a bottleneck dimension of $512$. 
 We otherwise use the same hyperparameters as \citet{Tang2020MultilingualTW} and report them in the Appendix.

\begin{table*}[ht]
\resizebox{\textwidth}{!}{%
\centering
\small

\begin{tabular}{lrrrrrrrrrrrrrrrrrr}
\toprule
   \multirow{3}{*}{Model}& \multicolumn{9}{c}{\textsc{Balto-}} &  \multicolumn{3}{c}{\textsc{Austro-}} &  \multicolumn{5}{c}{\textsc{Indo-}} &   \\ 
& \multicolumn{9}{c}{\textsc{Slavic}} &  \multicolumn{3}{c}{\textsc{nesian}} &  \multicolumn{5}{c}{\textsc{Iranian}} &   \\ \cmidrule(rl){2-10} \cmidrule(rl){11-13} \cmidrule(rl){14-18}  
&  bg$^\star$ & sr$^\star$ & hr & uk & sk$^\star$ & mk & sl & bs$^\star$ & be$^\star$ &  id & ms$^\star$ & fil$^\star$ & fa & hi & mr & ku$^\star$ & bn  & AVG \\ 

\midrule
\textbf{OPUS-100}  & &  & & &  &  &  & &   & & & & & & &  \\
Lang-pair  & \textbf{27.8} & 17.5 & 23.7 & \textbf{17.7} & 25.0 & \textbf{35.0} & \textbf{24.1} & \textbf{21.0} & 10.1 & 28.0 & 24.5 & - & \textbf{10.5} & 15.6 & 17.0 & 14.1 & 13.0 & 20.3 \\
Lang-agnostic  & 21.6 & 19.7 & 21.4 & 13.8 & 24.1 & 28.9 & 19.6 & 19.5 & 11.3 & 28.6 & 21.8 & - & 8.1 & 16.9 & 17.8 & 12.8 & 11.2 & 18.6\\ 
Lang-family  & 25.4 & \textbf{20.9}  & 23.7 & 15.1 & \textbf{27.7} & 31.9 & 22.6 &20.3 & \textbf{15.2}& \textbf{31.3} & \textbf{25.4} & - & 9.8 & \textbf{18.7} & \textbf{25.0} & \textbf{15.3} & 12.9 & \textbf{21.3}  
 
\\ \midrule 
\textbf{TED} & &  & & &  &  &  & &   & & & & & & &  \\
 Lang-pair & \textbf{35.7} & 21.1 & 30.5 & 21.1 & \textbf{24.2} & 27.0 & 21.4 & \textbf{28.6} & \textbf{12.5} & \textbf{35.4} & \textbf{23.4 }& \textbf{12.2}  & 14.0 & 14.1 & 10.0 & \textbf{4.9} & 9.0 & 20.3  \\
 Lang-agnostic & 31.7 & 24.0 & 29.7 & 21.9 & 20.6 & 26.5 & 20.2 & 27.8 & 7.7 & 33.8 & 22.1 & 11.6 & 17.0 & 15.5 & 7.0 & 3.3 & 6.0 & 19.2   \\
 Lang-family & 33.8 & \textbf{25.1} & 30.5 & 22.2 & 22.8 & \textbf{28.0} & 21.5 & 27.8 & 9.5 & 34.7 & 22.0 & 11.5 & \textbf{17.5} & \textbf{19.8} & 10.3& 4.1 & \textbf{11.6} & \textbf{20.7}  \\

\bottomrule

\end{tabular}}
\caption{Test set $\bleu$ scores when translating out of English (\textit{en} $\rightarrow$ \textit{xx}) on OPUS-100 and TED. 
$\langpair$ stands for language-pair, $\langagnostic$ for language-agnostic, and $\langfamily$ for language-family adapters. Languages denoted with $\star$ are new to mBART-50.  
Results in bold are significantly different (p < $0.01$) from the best adapter baseline.
}
\label{table:results}
\end{table*}

\section{Results and Discussion}

 \subsection{Main results} Table \ref{table:results} shows translation results for a subset of languages of OPUS-100 and TED in terms of $\bleu$ using parallel data to fine-tune mBART-50 in the $en\rightarrow xx$ direction. We also report $\comet$ scores in the Appendix.

Our approach ($\langfamily$) consistently improves results on the OPUS-100 dataset, with an average $+1$ $\bleu$ performance boost 
across all languages compared to fine-tuning with $\langpair$ adapters  and $+2.7$ improvement compared to $\langagnostic$ adapters. We believe that this shows that representations from similar languages are beneficial to a multilingual model in a low-resource setup. However, training a single adapter over all languages ($\langagnostic$) is detrimental in terms of translation quality. Moreover, $\langpair$ trains a different adapter on each language pair and does not permit sharing cross-lingual information. As a result, it obtains worse results compared to our approach; it is also significantly more computationally expensive, requiring $5\times$ parameters of $\langfamily$ adapters. 

\begin{table}[t] 
\resizebox{\columnwidth}{!}{%
\centering 
\small
\begin{tabular}{lrrr} 
     \toprule      
     &       \textbf{Parameters} & \textbf{Runtime} & \textbf{GPUs}\\   
    \midrule
 \textbf{$\langagnostic$} &  27M    & 35h & 8 \\
  \textbf{$\langfamily$}    &  81M   & 78h & 8 \\
  \textbf{$\langpair$}   & 432M  & 192h & 8 \\ 
  \textbf{$\mlft$}  & 680M & 310h & 8 \\
  \bottomrule      
\end{tabular}}
	\caption{Parameters used by our approach and the baselines to train on OPUS-100. We note that the GPUs used are NVIDIA-V100. For completeness, we also include the parameters used for multilingual fine-tuning ($\mlft$) of the pre-trained model. }

	\label{table:parameters}
\end{table}

 Our approach similarly outperforms both baselines on TED. It yields a $+1.5$ improvement compared to $\langagnostic$ and $+0.4$ BLEU compared to $\langpair$. These results confirm our main finding, which is that selectively sharing parameters of related languages with adapters is useful for low-resource NMT. 
 
 \subsection{Computational cost} We show in Table \ref{table:parameters} the number of trainable parameters used for each approach. We note that our experiments were conducted using $8$ NVIDIA-V100 GPUs. The mBART-50 model has $680$M parameters. Our approach trains parameters that add up to just $11.9\%$ of the full model. $\langagnostic$ is the most efficient approach, requiring just $8.4\%$ trainable parameters. However, there is a cost in terms of performance compared to our model. Finally, training $\langpair$ adapters is relatively expensive ($52.2\%$ of the trainable parameters of mBART-50). All in all, our $\langfamily$ approach provides a trade-off between performance and efficiency in terms of model parameters and is an effective method of adapting pretrained multilingual models to  low-resource languages. 

\begin{table*}[t]
\resizebox{0.96\textwidth}{!}{%
\centering
\small

\begin{tabular}{lrrrrrrrrrr}
\toprule
    &  \multicolumn{4}{c}{\textsc{Balto-}} &   \multicolumn{2}{c}{\textsc{Austro-}} &  \multicolumn{3}{c}{\textsc{Indo-}} &    \\ 
        &  \multicolumn{4}{c}{\textsc{Slavic}} &   \multicolumn{2}{c}{\textsc{nesian}} &  \multicolumn{3}{c}{\textsc{Iranian}} &    \\ \cmidrule(rl){2-5} \cmidrule(rl){6-7} \cmidrule(rl){8-10}
&  bg & hr & mk & be &  id & ms & fa & ku& bn  & AVG-16  \\ \midrule
$\langagnostic$ w/o emb adapter & 21.3 & 21.5 & 28.3 & 10.5 & 28.7 & 21.5 & 7.6 & 12.4 & 10.9 & 18.1 \\
$\langagnostic$ with emb adapter (\textsc{baseline}) & 21.6 & 21.4 & 28.9 & 11.3& 28.6 & 21.8 & 8.1 & 12.8 & 11.2 & \textbf{18.6} \\
$\langfamily$ w/o emb adapter & 24.3 & 22.6 & 31.2 & 13.4 & 31.4 & 25.2 & 9.0 & 13.7 &  12.2& 20.6 \\
$\langfamily$ with emb adapter (\textsc{ours}) & 25.4 & 23.7 & 31.9 & 15.2 & 31.3 & 25.4 & 9.8 & 15.3 & 12.9 & \textbf{21.3} \\\bottomrule
\end{tabular}}

\caption{Ablation of the proposed architecture for $en\rightarrow xx$ ($\bleu$ scores) on OPUS-100. We present results only for a subset of languages per language family. Full results can be found in the Appendix. }
\label{table:ablation}
\end{table*}

\subsection{Embedding-layer adapter}  Our approach keeps the encoder and decoder embeddings frozen during fine-tuning.
   Because of that, the lexical representations of the model are not updated to model the languages of interest. To overcome this issue, we introduce an adapter after the \textit{encoder embedding layer}, as well as after the \textit{decoder embedding layer}. We do not tie these adapter layers, since they only add up a small number of parameters ($1$M each, i.e., $0.1\%$ of mBART-50 parameters). 
   
   As we can see in Table \ref{table:ablation}, we get  consistent gains across almost all language pairs by adding these adapters, for both our model and the $\langagnostic$ baseline. The former yields a $+0.5$ performance boost, while the latter a $+0.7$ improvement in terms of BLEU. While the gains are modest, they are consistent and come at a very small computational overhead. For some languages, such as Kurdish (which is an unseen language for mBART-50), results improve by $+1.6$ when using embedding-layer adapters. Since Kurdish is not part of mBART-50 pretraining corpus, encoding token-level representations is in this case more challenging and embedding-layer adapters allows the model to specialize in this language.

\subsection{Automatic clustering of languages}
\noindent \textbf{Gaussian Mixture Model.} For our main set of experiments, we used language families from WALS. However, it might be that not all languages within a language family share the same linguistic properties \cite{ahmad-etal-2019-difficulties}. 
Therefore, we wanted to explore a data-driven approach to induce similarities between languages. To this end, we group languages together using Gaussian Mixture Model (GMM) clustering of text representations obtained from a PLM \cite{aharoni-goldberg-2020-unsupervised}. We used released code by the authors of the paper.\footnote{\url{https://github.com/roeeaharoni/unsupervised-domain-clusters}}

\begin{table*}[h]
\resizebox{\textwidth}{!}{%
\centering
\small

\begin{tabular}{lrrrrrrr}
\toprule
& \multicolumn{3}{c}{Language Groups} & id & fa & ku & AVG \\ \midrule
ling. family (ours) & <be, bg, sr, hr, uk, sk, mk, sl, bs> & <id, ms> & <ku, fa, hi, mr,  bn> &  31.3 & 9.8 & 15.3 & 21.3 \\
GMM& <bg, sr, hr, uk, sk, mk, sl, bs> & <\textbf{ku}, id, ms> & <\textbf{be}, fa, hi, mr,  bn>  & 29.7 & 9.2 & 14.3 & 19.4  \\
random & <bg, hr, mk, bs, be, ms, hi, mr, ku> & <sl, id> & <sr, uk, sk, fa, bn> &  27.8 &  7.0 & 15.0 &  18.4 \\ \bottomrule
\end{tabular}}

\caption{Evaluation of different methods to form language families for $en\rightarrow xx$ on OPUS-100. We present results only for a subset of languages and the overall average $\bleu$ scores. Full results are shown in the Appendix. }
\label{table:grouping}
\end{table*}
We use XLM-R \cite{conneau-etal-2020-unsupervised}, a multilingual PLM and specifically the  \textit{xlmr-roberta-base} HuggingFace \cite{wolf-etal-2020-transformers} checkpoint. We encode 500 sequences of 512 tokens from each language (using OPUS-100) to create sentence representations, by performing average pooling of the last hidden state. We then use PCA projection of dimension 100 and fit the sentence representations to a GMM with 3 components (3 Gaussian distributions, i.e., clusters). As this is a soft assignment, every language belongs with some probability to one or more clusters. For simplicity, we map each language to just one cluster based on where the majority of its samples are assigned to. 

\noindent \textbf{Results.} Table \ref{table:grouping} shows an evaluation of our approach, where we select the language family based on linguistic similarities (\textit{ling. family}, first row), GMM clustering (second row), and random sampling (third row). 

The main observation is that training adapters using language groups computed by GMM clustering yields worse translation scores compared to language groups based on linguistic similarities (\textit{ling. family}). We believe that this is the case because some languages were clustered together with linguistically distant languages (e.g., Belarusian is assigned to the same group as Persian, Hindi, Marathi, and Bengali according to GMM clustering). This might be because of a domain mismatch between the English-Belarusian parallel dataset and the datasets of the rest of the languages in the group. Based on our experiments, training adapters on linguistic families provides better translation scores and should therefore be preferred, if these exist. 
As expected, randomly clustering languages together performs worse than all approaches, showing that taking into account similarities between languages is beneficial when training a multilingual model for low-resource NMT. 

\section{Analysis} 

\subsection{Performance according to language family}
\begin{figure}
	\centering
	\includegraphics[width=1.\columnwidth, page=1]{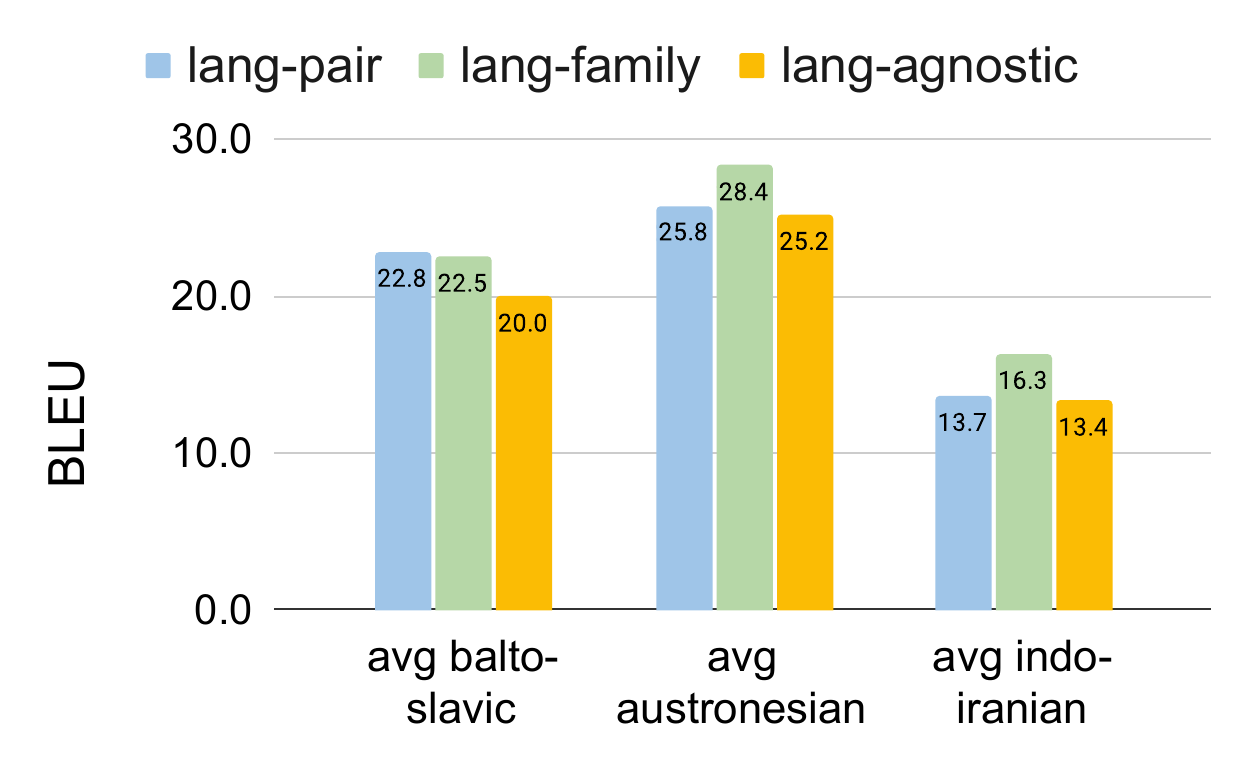}
	\caption{Grouping based on language family using OPUS-100. Translation scores (measured with BLEU) are shown for the our method ($\langfamily$), as well as the $\langpair$ and $\langagnostic$ baselines.}
	\label{fig:avg_per_lang_fam}
\end{figure}

To evaluate the contribution of grouping languages based on linguistic information, we present the $\bleu$ scores of the $\langfamily$ adapters compared to the baselines \textit{per language family}. We show the results in Figure \ref{fig:avg_per_lang_fam}. 

Compared to the $\langagnostic$ baseline, $\langfamily$ adapters perform better in all language families. On Balto-Slavic, our approach is on par with  $\langpair$ adapters (<0.5  $\bleu$ difference). On both Austronesian and Indo-Iranian, our approach largely outperforms (more than +2 $\bleu$) both baselines. This is arguably the case because $\langagnostic$ adapters, trained using parallel data from all languages, group dissimilar languages together and do not take into account language variation. We instead train adapters on languages with common linguistic properties and obtain consistently improved translations. 

$\langagnostic$ adapters perform worse than $\langpair$ adapters on all  language families. This is mostly evident for Balto-Slavic. We believe that this happens because Balto-Slavic languages are more similar to English compared to Austronesian or Indo-Iranian. This means that translating between Balto-Slavic and English is relatively easier, especially since mBART-50 has been trained with a large Indo-European bias and it already encodes cross-lingual information for most of the languages in this group. As a result, $\langpair$ adapters create in this case a very competitive baseline. 

\subsection{Performance on seen \textit{vs} unseen languages}
\begin{figure}[t]
	\centering
	\includegraphics[width=1.\columnwidth, page=1]{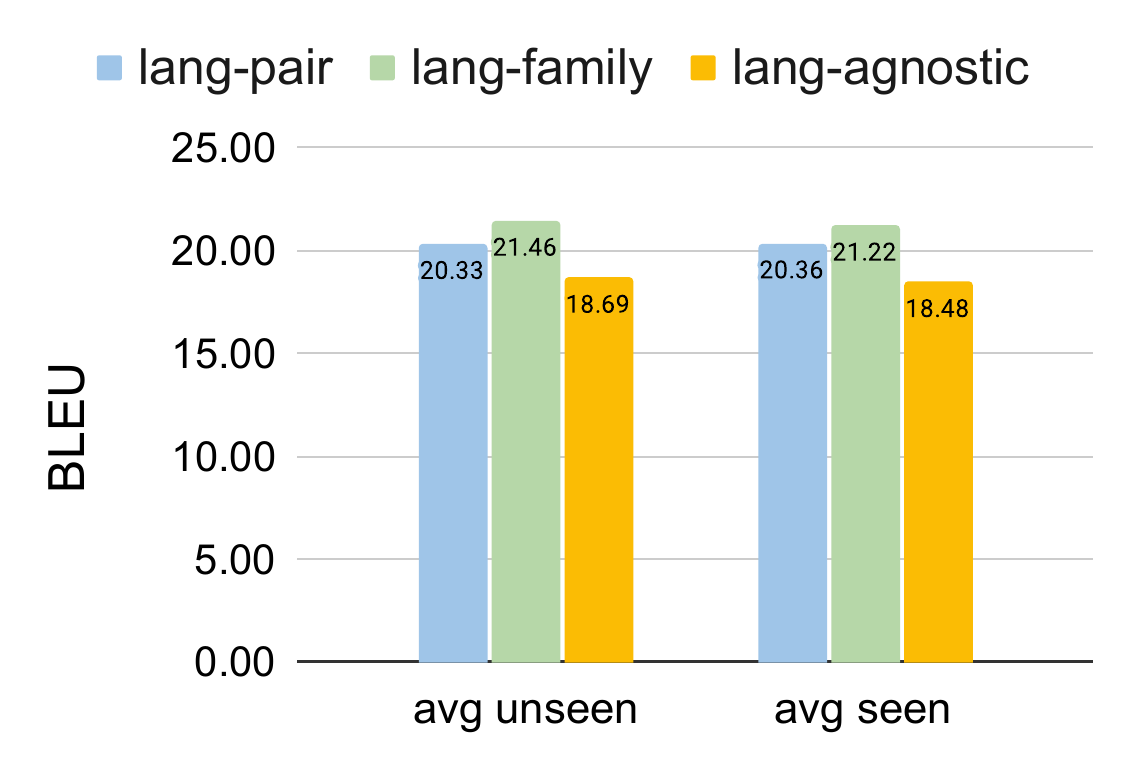}
	\caption{Grouping based on ``seen'' (existing in the pretraining corpus), or ``unseen'' language using OPUS-100. BLEU scores are shown for our method ($\langfamily$) and the baselines.}
	\label{fig:avg_per_seen_unseen}
\end{figure}

 We also evaluate the performance of language-family adapters and the baselines on languages that are not included in the mBART-50 pretraining data (\textit{unseen}), compared to languages that belong to its pretraining corpus (\textit{seen}). We present the results in Figure  \ref{fig:avg_per_seen_unseen}. 
 
 On unseen languages,  $\langfamily$ adapters improve the translation quality compared to the $\langpair$ adapter baseline. As the pretrained model has no knowledge of these languages, $\langfamily$ adapters provide useful cross-lingual signal. This makes our approach suitable for extending an already trained multilingual model to new languages in a scalable way. The improvement is, as expected, smaller for the seen languages.

$\langagnostic$ adapters perform significantly worse than both our approach and the $\langpair$ baseline. This might be the case because of negative transfer between unrelated languages, that are clustered and trained together using the $\langagnostic$ model. This issue is prevalent for both seen and unseen languages.

\section{Conclusion}
We presented a novel approach for fine-tuning a pretrained multilingual model for NMT using language-family adapters. Our approach can be used for low-resource multilingual NMT, combining the modularity of adapters with effective cross-lingual transfer between related languages. We showed that language-family adapters perform better than both language-agnostic and language-pair adapters, while being computationally efficient. 
Finally, for languages new to mBART-50, we showed that our approach provides an effective way of leveraging shared cross-lingual information between similar languages, considerably improving translations
compared to the baselines.

In the future, a more elaborate approach to encode lexical-level representations could further boost the performance of language-family adapters. We
also
hypothesize that the effectiveness of our model could be leveraged for other cross-lingual tasks, such as natural language inference, document classification and question-answering.

\section*{Limitations}

Our work uses a large seq2seq multilingual pretrained model, mBART-50. This model has been pretrained on large chunks of monolingual data from Common Crawl \cite{wenzek-etal-2020-ccnet}, but we do not have evaluations of generated text (e.g., on fluency, factuality, or other common metrics used to evaluate generated language). Therefore, this pretrained model can encode biases that could harm marginalized populations \citep{stochastic-parrots} and could also be used to translate harmful text. 

\section*{Acknowledgements}

 This project has received funding from the European Research Council under the European Union’s Horizon $2020$ research and innovation program  (grant agreement 
\#$640550$) and from DFG (grant FR $2829$/$4$-$1$). 

\bibliography{anthology,custom}
\bibliographystyle{acl_natbib}


\appendix

\section{Appendix}

\subsection{Dataset statistics}
\label{ssec:appendix2}

First, we show the script and language family (according to linguistic information) of each language used in our set of experiments in Table \ref{table:data}. We also present in detail the statistics of all parallel data used in our set of experiments in Table \ref{table:datasets}. We note that the number of train, validation and test set presented  refers to sentences. 

The TED dataset can be downloaded 
from \href{http://phontron.com/data/ted_talks.tar.gz}{phontron.com/data/ted\_talks.tar.gz} while OPUS-100 can be downloaded from \href{https://object.pouta.csc.fi/OPUS-100/v1.0/opus-100-corpus-v1.0.tar.gz}{object.pouta.csc.fi/OPUS-100/v1.0/opus-100-corpus-v1.0.tar.gz}. 
\begin{table}[h]
\centering
\small
\resizebox{\columnwidth}{!}{
\begin{tabular}{lrr}
\toprule
Language (code) & Family   & Script  \\  \midrule
$\star$Bulgarian (bg) & Balto-Slavic &  Cyrillic \\
\hspace{1.6mm}Persian (fa) & Indo-Iranian &  Arabic \\
$\star$Serbian (sr) & Balto-Slavic  & Cyrillic  \\
\hspace{1.6mm}Croatian (hr) & Balto-Slavic  &Latin \\
\hspace{1.6mm}Ukrainian (uk) & Balto-Slavic & Cyrillic \\
\hspace{1.6mm}Indonesian (id) & Austronesian  &  Latin  \\
$\star$Slovak (sk) & Balto-Slavic  & Latin \\
\hspace{1.6mm}Macedonian (mk) & Balto-Slavic &Cyrillic  \\
\hspace{1.6mm}Slovenian (sl) & Balto-Slavic  &  Latin  \\
\hspace{1.6mm}Hindi (hi) & Indo-Iranian   & Devanagari \\
\hspace{1.6mm}Marathi (mr) & Indo-Iranian & Devanagari  \\
$\star$Kurdish (ku) & Indo-Iranian &  Arabic  \\
$\star$Bosnian (bs) & Balto-Slavic  &  Cyrillic \\
$\star$Malay (ms) & Austronesian  &  Latin  \\
\hspace{1.6mm}Bengali (bn) & Indo-Iranian  &  Bengali  \\
$\star$Belarusian (be) & Balto-Slavic &  Cyrillic  \\
$\star$Filipino (fil)  & Austronesian &  Latin  \\

\bottomrule

\end{tabular}}
\caption{Languages that are used in the experiments. $\star$ indicates languages that are \textit{unseen} from mBART-50, i.e., they do not belong to the pretraining corpus. Filipino is only used in the TED experiments.}

\label{table:data}
\end{table}

\begin{table}[ht]
\resizebox{\columnwidth}{!}{%
\centering
\small
\begin{tabular}{lrrr}
\toprule
 Adapter size & Dropout & Lang-Family & Lang-Agnostic\\ \midrule
 128 & 0.1 & 16.8 & 10.1 \\
128& 0.3 & 16.4 & 9.5 \\
 256 & 0.1 & 19.0 & 14.9  \\
 256 & 0.3 & 18.6 & 14.0 \\
 512 & 0.1 & 20.7 & 19.2  \\
  512 & 0.3 & 19.9 & 18.5  \\ \bottomrule

\end{tabular}}
\caption{Hyperparameter tuning for dropout, adapter bottleneck size on TED. Average performance (on all language pairs using TED) per model. We chose the best-performing combination of dropout and bottleneck size for our experiments. }
\label{table:tuning}
\end{table}

\subsection{Training details}
\label{ssec:appendix_training}
We train each model for $130$k updates with a batch size of $900$ tokens per GPU for OPUS-100 and $1024$ tokens per GPU for TED. We use $8$ NVIDIA-V100 GPUs for OPUS-100 and $2$ GPUs for TED (much smaller dataset). We evaluate models after $5$k training steps. We use early stopping with a patience of 5. To balance high and low-resource language pairs, we use temperature-based sampling \cite{arivazhagan2019massively} with $T = 1.5$.

\subsection{Evaluation of main results using 2 metrics}
\label{ssec:appendix1}

We evaluate the translations of our model ($\langfamily$ adapters) and all the baselines trained on OPUS-100 using $\comet$ \cite{rei-etal-2020-comet}.
$\comet$ leverages progress in cross-lingual language modeling, creating a multilingual machine translation evaluation model that takes into account both the source input and a reference translation in the target language. We rely on \texttt{wmt-large-da-estimator-1719}. $\comet$ scores are not bounded between 0 and 1; higher scores signify better translations. Our results are summarized in Table \ref{table:results_ablatio}. We see that $\comet$ correlates with $\bleu$
in our experiments.

\subsection{Hyperparameters}
\label{ssec:appendix3}
 We tune the dropout and the adapter bottleneck size on TED. We use values {0.1, 0.3} for the dropout and {128, 256, 512} for the bottleneck size. We list the hyperparameters we used to train both our proposed model and the baselines in Table \ref{table:hyperparameters}.

\begin{table*}[ht]
\centering
\small

\begin{tabular}{lrrrr|rrrr}
\toprule

\textbf{Language} & \textbf{Source}  & \textbf{Train} & \textbf{Valid} & \textbf{Test} & \textbf{Source} & \textbf{Train} &\textbf{Valid} & \textbf{Test}\\  \midrule 
Bulgarian (bg) & TED  &  174k & 4082 & 5060 & OPUS-100 &1M & 2k & 2k \\
Persian (fa) & TED & 151k &3930 & 4490&OPUS-100& 1M& 2k  & 2k \\
Serbian (sr) & TED & 137k & 3798 & 4634 &OPUS-100& 1M & 2k  & 2k \\
Croatian (hr) & TED & 122k & 3333 & 4881 &OPUS-100& 1M& 2k  & 2k\\
Ukrainian (uk) & TED & 108k & 3060 & 3751&OPUS-100&1M & 2k  & 2k\\
Indonesian (id) & TED  & 87k& 2677 & 3179 &OPUS-100&1M & 2k  & 2k\\
Slovak (sk) & TED  & 61k & 2271 & 2445 &OPUS-100&1M& 2k  & 2k\\
Macedonian (mk) & TED & 25k  & 640 & 438&OPUS-100&1M& 2k  & 2k\\
Slovenian (sl) & TED  & 20k & 1068 & 1251&OPUS-100&1M& 2k  & 2k\\
Hindi (hi) & TED & 19k & 854 & 1243 &OPUS-100&534k& 2k  & 2k\\
Marathi (mr) & TED & 10k & 767 & 1090 &OPUS-100&27k& 2k  & 2k\\
Kurdish (ku) & TED & 10k& 265 & 766 &OPUS-100&45k& 2k  & 2k\\
Bosnian (bs) & TED  & 6k & 474 & 463 &OPUS-100&1M & 2k  & 2k\\
Malay (ms) & TED  & 5k & 539 & 260  &OPUS-100&1M& 2k  & 2k\\
Bengali (bn) & TED  & 5k & 896 & 216 &OPUS-100&1M & 2k  & 2k\\
Belarusian (be) & TED & 5k & 248 & 664 &OPUS-100&67k& 2k  & 2k\\
Filipino (fil)  & TED2020 & 3k & 338 & 338&OPUS-100& - & - & - \\

\bottomrule

\end{tabular}
\caption{Dataset details for TED \cite{qi-etal-2018-pre,reimers-2020-multilingual-sentence-bert} and OPUS-100 \cite{zhang-etal-2020-improving}.}
\label{table:datasets}
\end{table*}

\begin{table}[t]
\centering
\small

\begin{tabular}{lr}
\toprule

\textbf{Hyperparameter} & Value \\  \midrule 
Checkpoint & mbart50.pretrained \\ 
Architecture & mbart\_large \\
Optimizer & Adam \\ 
$\beta_{1}, \beta_{2}$ & 0.9, 0.98 \\
Weight decay & 0.0 \\
Label smoothing & 0.2 \\ 
Dropout & 0.1 \\
Attention dropout & 0.1 \\ 
Batch size & 1024 tokens \\ 
Update frequency & 2 \\ 
Warmup updates & 4k \\ 
Total number of updates & 130k\\ 
Max learning rate & 1e-04 \\ 
Temperature sampling & 5\\ 
Adapter dim. & 512\\ 
\bottomrule

\end{tabular}
\caption{Fairseq hyperparameters used for our set of experiments.}
\label{table:hyperparameters}
\end{table}

\begin{table*}[h]
\centering
\small

\begin{tabular}{lrrrrrrrrr} \toprule
 &\multicolumn{2}{c}{\textbf{\langfamily}} & \multicolumn{2}{c}{\textbf{\langpair}} &  \multicolumn{2}{c}{\textbf{\langagnostic}} &  \multicolumn{2}{c}{\textbf{\mlft}} \\ 
  \textbf{Lang} & $\bleu$ & \comet  & $\bleu$   & \comet   & $\bleu$ & \comet & $\bleu$   & \comet \\ 
  \cmidrule(rl){2-3} \cmidrule(rl){4-5}   \cmidrule(rl){6-7}   \cmidrule(rl){8-9} 

bg & 25.4 & 67.2 & 27.8 & 72.1 & 21.6 & 44.6 & 28.0&  76.5   \\
sr & 20.9& 44.3 & 17.5  & 38.2 & 19.7 & 41.1 & 21.1&  48.4 \\
hr & 23.7 & 55.0 & 23.7  & 53.1 & 21.4 & 43.4 &24.5   & 55.1   \\
uk & 15.1 & -17.0 & 17.7 & 14.4  & 13.8 & -18.5 & 17.1   & 35.9   \\
sk & 27.7& 54.3 & 25.0 & 50.1     & 24.1 & 57.0 &  30.5   & 64.9   \\
mk & 31.9& 62.9 & 35.0 & 64.1     & 28.9 & 65.2 & 35.6   & 62.1    \\
sl & 22.6& 48.9   & 24.1 & 65.8   & 19.6 & 42.3 & 24.5& 64.3\\
bs & 20.3& 44.1  & 21.0 & 37.1 & 19.5 & 43.9 & 22.1   &  50.8  \\
be & 15.2& -10.2   & 10.1 & -21.6 & 11.3 & -13.9 & 17.9   &  36.6   \\
id & 31.3& 60.1  & 28.0 & 64.0  & 28.6 & 77.0 & 31.5   & 60.1   \\
ms & 25.4 & 53.5 & 24.5 & 66.1  & 21.8 & 49.8  & 25.5   & 68.0   \\
fa & 9.8 & -23.5  & 10.5 & -22.1& 8.1 & -24.4 & 9.5 & -15.0     \\
hi & 18.7& 39.1  & 15.6 & -19.1 & 16.9 & 10.1 & 18.4 & 36.4   \\
mr & 25.0& 67.0  & 17.0 & 9.0  & 17.8 & 19.5 & 24.7 & 58.1    \\
ku & 15.3& -18.5  & 14.1 &  -12.9 & 12.8 & -11.5 &15.6 & -9.1     \\
bn & 12.9&  -16.0  & 13.0 & -24.1 & 11.2 & -18.1 & 14.1 & -8.5    \\ \midrule 
avg & 21.3 & 32.0 & 20.3 & 27.1  & 18.6 &25.5 & 22.5 & 42.8  \\
\bottomrule

\end{tabular}
\caption{Test set $\bleu$ and $\comet$ scores when translating out of English using OPUS-100. Languages are presented by decreasing amount of parallel data per language family. $\langpair$ stands for language-pair adapters, $\langagnostic$ for language-agnostic, while $\langfamily$ for language-family adapters. $\mlft$ stands for multilingual fine-tuning of the entire mBART-50 model.}
\label{table:results_ablatio}
\end{table*}

\begin{table*}[t]
\resizebox{0.96\textwidth}{!}{%
\centering
\small

\begin{tabular}{lrrrrrrrrrrrrrrrrr}
\toprule

&  bg$^\star$ & sr$^\star$ & hr & uk & sk$^\star$ & mk & sl & bs$^\star$ & be$^\star$ &  id & ms$^\star$ & fa & hi & mr & ku$^\star$ & bn  & AVG \\ \midrule 

Lang-agnostic w/o emb  & 21.3 & 19.0 & 21.5 & 13.9 & 23.6 & 28.3  & 19.1 & 18.9& 10.5 & 28.7 & 21.5 & 7.6  & 16.1& 16.9 & 12.4 & 10.9 & 18.1 \\
Lang-agnostic with emb  & 21.6 & 19.7 & 21.4 & 13.8 & 24.1 & 28.9 & 19.6 & 19.5 & 11.3  & 28.6& 21.8 & 8.1& 16.9 & 17.8 & 12.8 & 11.2 & 18.6  \\
Lang-family w/o emb & 24.3 & 20.4 & 22.6 & 14.8 & 26.3 & 31.2 & 21.9 & 20.6 & 13.4& 31.4 & 25.2 & 9.0 & 18.3 & 23.7 & 13.7 & 12.2 & 20.6 \\
Lang-family with emb & 25.4 & 20.9 & 23.7 & 15.1 & 27.7 & 31.9 & 22.6 & 20.3 & 15.2 & 31.3 & 25.4 & 9.8 & 18.7 & 25.0 & 15.3 & 12.9 & \textbf{21.3} \\ \bottomrule
\end{tabular}}

\caption{Full results of the ablation of the proposed architecture for $en\rightarrow xx$ ($\bleu$ scores) on OPUS-100. Bold results indicate best performance on average. }
\label{table:ablation_full}
\end{table*}

\subsection{Embedding-layer results}
\label{ssec:appendix4}

We report in Table \ref{table:ablation_full} the results of the ablation study concerning the use of \textit{embedding-layer} adapters on all languages.

\subsection{Results using GMM, random clustering and language families}
Full results of Table \ref{table:grouping} can be seen in Table \ref{table:grouping2}.

\begin{table*}[h]
\resizebox{\textwidth}{!}{%
\centering
\small

\begin{tabular}{lrrrrrrrrrrrrrrrrrr}
\toprule
&  bg & sr & hr & uk & sk & mk & sl & bs & be &  id & ms & fil & fa & hi & mr & ku & bn  & AVG  \\ \midrule
GMM & 23.9 & 17.7 & 24.4 & 11.0 & 19.3 & 22.9 & 19.0 & 23.6 & 14.9 & 29.7 & 23.4 & - & 9.2 & 18.8 & 25.5 & 14.3 & 13.2 & 19.4 \\
random & 22.9 & 18.8 & 23.5 & 10.0 & 22.5 & 31.9 & 21.1 & 20.1 & 12.1 & 25.8 & 24.9 & - &5.0& 18.6 & 22.9 & 15.0 & 8.1 & 18.4\\

\bottomrule
\end{tabular}}

\caption{Evaluation of different methods to form language families for $en\rightarrow xx$ ($\bleu$) on OPUS-100.  }
\label{table:grouping2}
\end{table*}

\end{document}